# Paper Manuscript

## Title

Deep learning-based hierarchical insect classification using camera trap imagery

## Authors

Zaki Mahfoud [a], Juan A. Chiavassa [b], Simon Walther [c], Florian Haselbeck [a],
Ehsan Yaghoubi [a]

## Affiliations

[a] Weihenstephan-Triesdorf University of Applied Sciences, Department of Sustainable Agriculture and Energy Systems, Professorship of Smart Farming, Am Staudengarten 1, 85354 Freising, Germany
[b] Weihenstephan-Triesdorf University of Applied Sciences, Institute for Biomass Research, Neuseser Str. 1, 91732 Merkendorf, Germany
[c] Weihenstephan-Triesdorf University of Applied Sciences, Department of Sustainable Agriculture and Energy Systems, Professorship of Digital Farm Management, Am Staudengarten 1, 85354 Freising, Germany

## Corresponding author

Ehsan Yaghoubi [a],
ehsan.yaghoubi@hswt.de

## Keywords

## Highlights

- manually-curated and pre-processed insect camera trap monitoring dataset from Germany.
- Deep learning-based hierarchical image classification leveraging biological taxonomy.
- Hierarchy maximises use of partially-labelled data, minimises unclassified samples.
- Strong performance on many rare classes due to class-balanced weighting.
- Limitations remain for low-sample, visually heterogeneous classes of small specimens.

# Abstract

Declining insect populations make reliable biodiversity monitoring increasingly urgent, yet monitoring of insect biodiversity is hampered by a lack of standardised data and by costly and time-consuming manual identification by expert entomologists. Deep learning-based image classifiers, processing data from automated non-lethal camera traps, have the potential to transform and scale insect biodiversity monitoring. However, challenges remain in acquiring expert-annotated datasets, developing model architectures that generalise well across diverse taxonomic levels and training models on highly imbalanced data. Hierarchical data also benefits from designing models that default to higher-confidence, coarser-level predictions, when uncertain about finer taxonomic levels. In this paper we address these challenges with a deep learning-based hierarchical classification model. First, we present an manually-curated, long-tailed dataset of around one million images of insects, extracted from 1,801 camera-trap video recordings and annotated with a five-level, 34-class hierarchy. Further, we adapt a hierarchical classification model architecture to a five-level variable-depth hierarchy, with class-balanced weighting. Our model improves on non-hierarchical classifiers by leveraging biological taxonomy to extract granularity-specific visual features and makes hierarchy-consistent predictions to the deepest taxonomic level that meets a confidence threshold ($T$ = 0.6). Our model achieved a per-level accuracy of 80-99% on test data, across five levels of hierarchy. Furthermore, our results show the model's ability to correctly classify (i) the parent class of wrongly predicted instances, (ii) rare classes in our highly imbalanced dataset, and (iii) visually similar, but taxonomically distinct, insects. Despite its limitations on very low-resolution images, our approach is a promising application of deep learning-based hierarchical classifiers to imbalanced datasets for biodiversity monitoring.

# 1 Introduction

Insects are a highly diverse class of animals that are of vital importance for both ecosystem-services and agriculture, as pollinators and natural pest controllers (Dainese et al., 2019; Ollerton, 2017), while also being agricultural pests of significant economic importance (Zalucki et al., 2012). Over the past several decades, arthropod populations have been reported to be in decline (Hallmann et al., 2017; Wagner et al., 2021), although population changes are not uniformly negative and vary both regionally and between taxa (Engelhardt et al., 2022; Van Klink et al., 2024). Changes in insect populations have implications not only for their diversity, abundance and biomass, but also for their function in ecosystems and their ability to adapt to future environmental change (Wildermuth et al., 2025). In this context, reliable and at-scale monitoring of insects is important to understand how their distributions are changing both spatially and temporally (Wagner et al., 2021). Despite this concerning trend, the monitoring of arthropod populations is hampered by a lack of (temporally and spatially) standardised data (Engelhardt et al., 2022) and of entomologists to carry out surveys (European Commission. Directorate General for Environment., 2022). Monitoring of insects is typically performed manually, which is time consuming and requires expert entomologists to identify species (Montgomery et al., 2021). Methods such as DNA Barcoding are often necessary to conclusively identify species with high intraspecific and low interspecific variation, but these are destructive and cannot provide information on the abundance of particular species (Høye et al., 2021).

Recent advances in deep learning (DL) methods have demonstrated their potential to transform biodiversity monitoring, providing a low-cost and scalable means of tracking changes in insect populations and diversity (van Klink et al., 2022). DL methods can be applied both to individual non-lethal monitoring devices and as part of a network of monitoring devices, providing real-time spatio-temporal data on biodiversity changes (Høye et al., 2021). DL-based image classifiers, e.g. convolutional neural networks (CNNs), trained on labelled images have been successfully employed for the monitoring of arthropods using automated camera traps (Bjerge et al., 2022; Sittinger et al., 2024; Tian et al., 2023). However, when considering the taxonomic diversity of insects, it is challenging to build a well-generalizing model without addressing the relationships between classes[1] and the granularity of labels (Chang et al., 2021; Deng et al., 2014). To generalise to many classes, a model must differentiate between classes that are mutually exclusive (e.g. a weevil cannot also be a grasshopper) and those that are not (a weevil is also a member of *Insecta* and *Coleoptera*). Hierarchical classification, where image labels can be structured according to a given hierarchy, for example, biological taxonomy for insects, addresses these challenges by encoding a semantic relationship between classes (Deng et al., 2014).

[1] The term ‘class’ is used in this paper as a general term for a labelled grouping of specimens at any taxonomic level, in addition to the taxonomic ‘class’ *Insecta* or *Arachnida*, used additionally in this paper.

It improves on flat classifiers by leveraging the hierarchical nature of biological taxonomy to extract level-specific visual features for coarse and fine classes (Chang et al., 2021; Elhamod et al., 2022). The effectiveness of hierarchical classifier models to share extracted features between levels and achieve high classification accuracy has been demonstrated on balanced datasets (equal numbers of images per class) of limited hierarchy depth (Chang et al., 2021; Chen et al., 2022). With respect to practical application, hierarchical classification provides the advantage of classifying at coarser levels of the taxonomy if a specific insect cannot be predicted with high enough confidence, e.g., since it was not included in the training data.

Training on real-world biological datasets, however, presents particular challenges, with highly imbalanced numbers of samples per class, low numbers of samples for minority classes and incomplete labelling to species level (Schneider et al., 2020). Developing DL-based models typically requires large datasets of labelled images to achieve acceptable performance, but acquiring and curating labelled datasets is costly and time-consuming (Høye et al., 2021). Camera traps themselves produce large amounts of data that can be used for training insect image classifiers (Bjerge et al., 2023a; Sittinger et al., 2024), but even in these cases challenges remain with obtaining adequate images for rare classes. The application of automated image classifiers in insect camera traps faces additional challenges at inference: Unlike a human taking a photograph of an insect to upload to an online platform for identification, an automated monitoring device cannot manoeuvre into an optimal position to view the specimen. It also does not know *a priori* which frame of a video feed will give the most confident and accurate classification. Here hierarchical classification has been demonstrated as a method to minimise uncertain predictions by classifying to the finest taxonomic level of the hierarchy that meets a threshold of confidence (Wu et al., 2020) and to tag species not seen during training (Bjerge et al., 2023b).

Despite this progress, a critical gap persists: existing hierarchical classification models require significant adaptations to work with variable-depth hierarchies, imbalanced datasets and images suffering from occlusions, suboptimal poses and varying lighting conditions. A model that can be easily adapted to a given hierarchy, be used with imbalanced and partially labelled data and suboptimal images would be very useful in the field. In this paper, we address this gap by proposing a DL-based computer vision model, employing a hierarchical classification approach, that addresses these challenges. First, we present a new manually-curated, long-tailed dataset of images of insects with around one million samples, extracted from 1,801 camera-trap video recordings, annotated with a five-level, 34-class taxonomic hierarchy and made available to the community to support future research. Second, we design a model architecture that makes hierarchy-aware predictions to the deepest taxonomic level that meets a user-defined confidence threshold in a variable-depth hierarchy. Third, we evaluate the model's performance systematically across all hierarchy levels and classes, including analysis of rare-class performance, visually confusable taxa, and the effect of the confidence threshold on prediction depth. The remainder of the paper is structured as follows: Section 2 reviews related work; Section 3 describes the dataset and model; Section 4 presents and discusses results; and Section 5 concludes.

# 2 Related work

In this section we review existing methods that leverage DL-based approaches for insect biodiversity monitoring, as well as works on hierarchical classification. In the course of this review we also consider the datasets used, in particular their acquisition, structure and imbalance.

## 2.1 Insect biodiversity monitoring

The use of DL-based computer vision models for field-based detection and classification of insects has been demonstrated by a number of studies (Bjerge et al., 2022; Sittinger et al., 2024; Stark et al., 2023; Tian et al., 2023). These systems vary in their taxonomic scope, dataset characteristics and classification architecture, with implications for their ability to generalise across taxa and deployment conditions. Examples of camera traps used in three of these papers are illustrated in Figure 1.

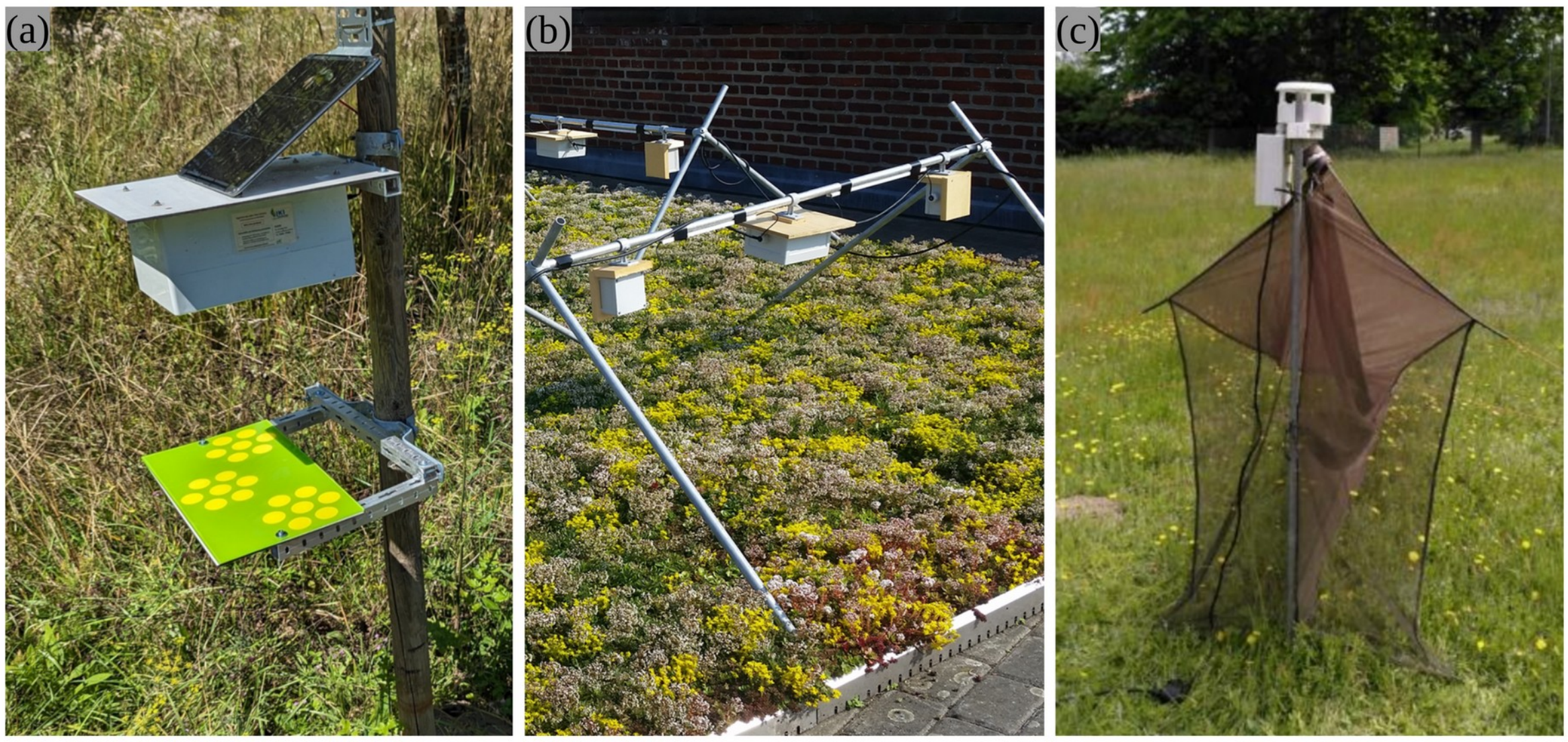

*Figure 1: Automated camera traps used for detecting and classifying insects. (a) Camera with platform from Sittinger et al., 2024; (b) Camera monitoring Sedum on a roof from Bjerge et al. 2023a; (c) 3D-printed camera trap attached to a Malaise trap net from Chiavassa et al. 2024.*

A straightforward implementation of these systems are automated agricultural pest traps. Typically targetting a limited number of species, they avoid both the need for hierarchical architectures and the challenge of class imbalance (Preti et al., 2021; Tian et al., 2023). For multi-class classifiers, insufficient labelled data for training, in particular for rare classes, as well as imbalanced datasets and image resolution for very small insects present challenges (Schneider et al., 2020; Teixeira et al., 2023). Schneider et al. (2020) found poor recall scores for classes trained on less than 500 images. The question of dealing with long-tailed species distributions during model training is compounded by the hierarchy: rare monotypic taxa remain under-represented at each level of the hierarchy. Cui et al. (2019) proposed a re-weighting method to correct for class imbalance during model training.

Sittinger et al. (2024) developed a camera trap (Figure 1, (a)) incorporating on-device detection and tracking and subsequent classification with a 27-class model. Even with relatively few classes, their flat architecture has an inherent limitation: treating parent and sub-classes as mutually exclusive, which we address in this paper. They used a model trained on a hand-curated dataset to automate the classification and creation of a much larger dataset from camera recordings. This reduces the annotation bottleneck and is directly relevant to the dataset construction approach adopted in this paper. Bjerge et al. (2023a) employed a similar process of training a classifier on a small, curated dataset and then using it to create a larger labelled dataset from time-lapse camera images using their camera trap (Figure 1, (b)). Their three-level (order, family, species) hierarchical insect classification model achieved high accuracy at all levels. While many studies use data labelled to species level, Stark et al. (2023) presented an eight-class model of arthropod orders to classify images of pollinators on flowers, demonstrating accurate predictions on coarse-grained classes that nevertheless provide useful data for biodiversity monitoring. Of particular relevance to the research in our paper is the Field Automatic Insect Recognition-Device (FAIR-Device), which uses a camera trap attached to a Malaise trap net (Figure 1, (c)) to detect and video-record insects passing through the device. This presents a non-lethal means of collecting close-up images of insects in the field for biodiversity monitoring (Chiavassa et al., 2024). Its use in a number of locations in 2021 and 2023 has created a dataset of over 2,500 videos of insects, from which the manually-curated dataset described in Section 3.1 was extracted.

## 2.2 Hierarchical classification

Semantically, hierarchical classification seeks to capture the relationships between class labels, to encode both mutually-exclusive and nested classes (Deng et al., 2014). In biological taxonomy, where relationships between classes are, to an extent, based on shared morphological features, the features of coarser taxa (e.g. class, order) can also inform the identification of finer taxa (e.g. family, genus, species) (Elhamod et al., 2022). Labelling images for training to species level requires sufficient domain knowledge, image resolution and visibility of distinguishing features (Chang et al., 2021; Chen et al., 2022; Wu et al., 2020). Hierarchical classifiers enable the use of images labelled at more than one taxonomic level in DL-based models. Hierarchical classification can also exploit the degree of relatedness between classes so that, even when a prediction is incorrect, it remains taxonomically close to the true label (Bertinetto et al., 2019). This requires a model architecture which encodes the degree of relation between classes, e.g. in the loss function. With this approach, misclassification between more closely related classes (for example, two genera in the same family) is less penalised than misclassification between distantly related classes.

Elhamod et al. (2022) presented a two-level (genus-species) hierarchy of fish samples that showed higher accuracy, especially for partially occluded images, when compared to a non-hierarchical (flat) model. However, the hierarchy is particularly shallow and is not enforced in the loss function. From a highly imbalanced raw dataset of museum samples, they created balanced subsets for training, validation and testing. Chang et al. (2021) proposed using three level-specific classifier heads to separate features by granularity, then concatenating the fine-grained features to improve coarse-grained predictions. Their implementation of independent cross-entropy loss on each level does not, however, encode the degree of relatedness between classes, nor is the effect of class imbalance addressed. Chen et al. (2022) expanded on the disentanglement of level-specific features presented in Chang et al. (2021) by implementing residual connections before the classification heads to add features from coarse-level classes (order, family) to fine-level classes (genus, species). Their hierarchical loss function encodes the relationships between taxa and makes hierarchy-consistent predictions. Thus, two species can be defined in the loss function as mutually-exclusive of one another, while both being in the same genus. Additionally, they model mixed-depth hierarchies (2- and 3-level) to simulate lack of domain knowledge, although the maximum depth is limited to three levels. While the architecture proves to be effective as a hierarchical classifier, results are based on a balanced dataset of birds, cars and aircraft, and do not address the class imbalance characteristic of real-world entomological data.

When a model can classify at coarser or finer levels, the confidence of its predictions at each level becomes an important design consideration. To this end Wu et al. (2020) proposed a method for classifying to the deepest class in a hierarchy that meets a threshold of confidence. While they propose a top-down approach and their applied confidence threshold is unspecified, we implement a bottom-up method, which we test at four probability thresholds. Another application of (low) probability thresholds was presented by Bjerge et al. (2023b), who implement anomaly detection in hierarchical insect classification model to tag novel taxa not seen in training. Wallin et al. (2025) proposed using probability distributions to predict out-of-distribution samples to the correct internal nodes in the hierarchy. Together, these works demonstrate the value of confidence-aware prediction depth as a mechanism for minimising uncertain classifications, which is a principle central to the model presented in this paper.

# 3 Materials and methods

In this section we present our dataset, including curation and hierarchical annotation. Then we describe our model architecture and loss functions, before detailing the experimental setup and consider the performance metrics we used.

## 3.1 Data

This section outlines the process of collection of the raw data, the extraction of the images used to create the manually-curated dataset as a contribution to this paper and their annotation for use in the hierarchical classification model. Figure 2 illustrates the workflow, from data collection, through extraction of cropped images and curation. All source code is available at https://github.com/smAIL-WS/HierarchicalInsectClassification and the dataset is available at 10.5281/zenodo.21690971.

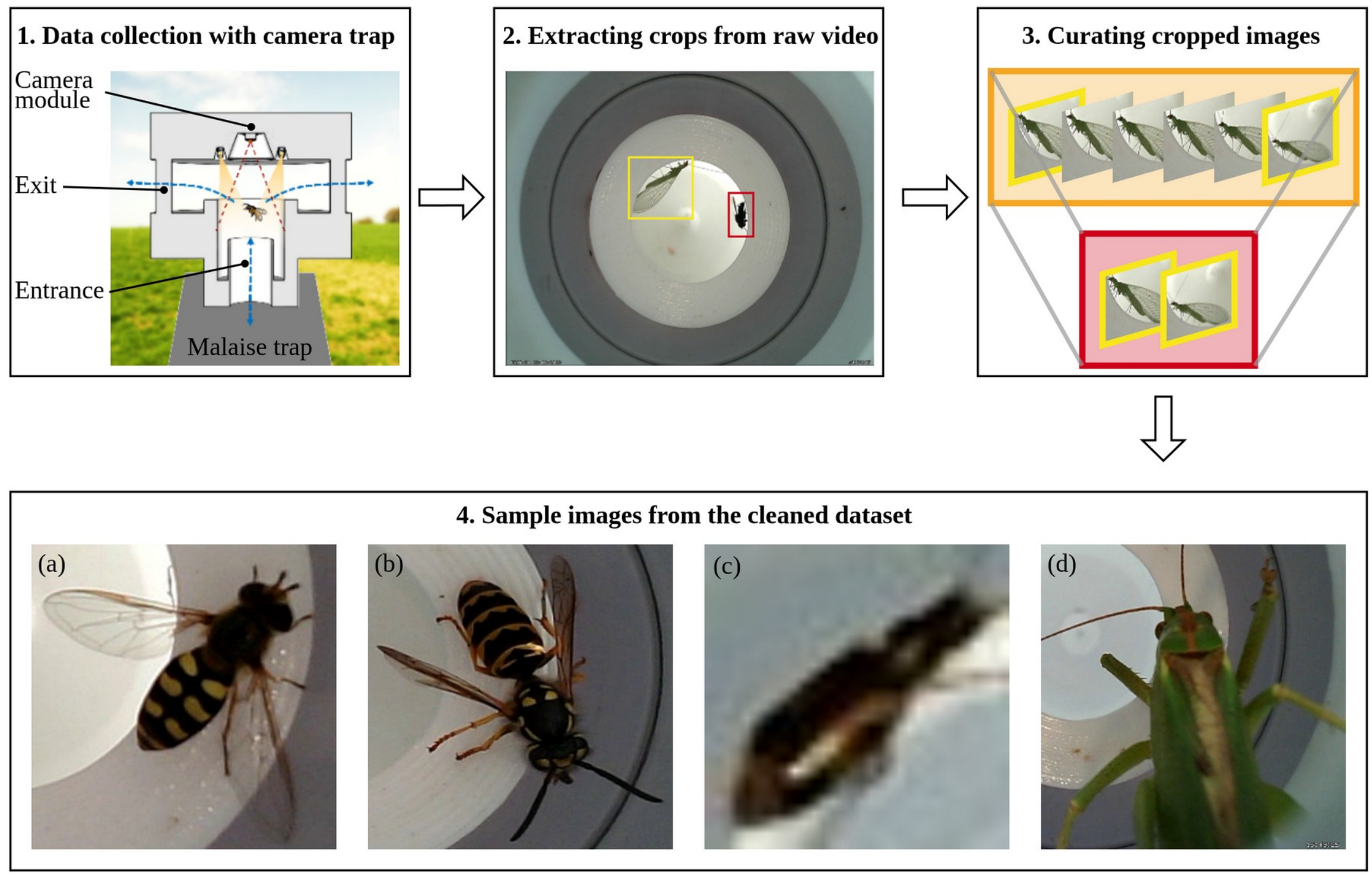

*Figure 2: The workflow comprises data collection (1), extraction of cropped images from individual frames of the video recordings (2) and curation of the cropped images (3). A sample of the cleaned dataset is presented in (4), showing visually similar Syrphidae (a) and Vespidae (b), low resolution samples (Anthicidae) (c) and specimens larger than the field-of-view (Orthoptera) (d).*

### 3.1.1 Data collection

Our starting point for this study was the raw video data collected and labelled in an earlier study with the so-called FAIR-Device automated camera trap. Further information on the trap design, and examples of data collected can be found in Chiavassa et al. (2024). The device replaces the typical lethal trap attached to the top of a Malaise-type insect trap and employs motion-activated video recording of insects (or other arthropods) passing through the apparatus. Video recordings were collected in July-August 2021 and August-September 2023 at four grassland sites in Lower Saxony, Germany and comprise 2,541 videos ranging from several seconds to several hours. The specimens in the videos were identified to varying granularity (from order to species level) using iNaturalist's (https://www.inaturalist.org/) DL-based and community-reviewed identification process. We used the raw dataset and annotations from Chiavassa et al. (2024) as the basis for our study, before preparing and processing the data, as outlined in the following section.

### 3.1.2 Data preparation

We first limited the source videos to those with a single labelled specimen, excluding videos with two or more specimens present ($n$ = 740), thereby reducing the dataset to 1,801 videos. This was done to reduce the instances of the wrong label being applied to the cropped images and thus to reduce the need for manual re-classification on this large dataset. The cropped still images were extracted from the video recordings using a YOLOv8-based object detector, pre-trained on invertebrates (https://github.com/jach999/fairtrack/). Each frame of the videos was processed to maximise the number of cropped images.

The extracted cropped images were curated to remove unlabelled specimens (including instances where two insects were present together) and erroneous crops (e.g. debris, empty frames). While some insects remain active while passing through the camera trap, thereby presenting very diverse poses that are beneficial for training, other insects remain motionless for extended periods which results in repetitive cropped images, despite representing only one individual. In the case of the latter the number of repetitive images was reduced by expertly curating the dataset to retain every $n^{th}$ image (where $n$ was determined per class based on the degree of repetition, ranging from 5 to 100).

We created a database including metadata for each sample, consisting of labels, image height and width and year of data collection. Images varied substantially in size and aspect ratio (see Figure S1 for full distributions by year). This heterogeneity necessitated preprocessing transformations (resizing, cropping, padding) to conform to model input requirements.

### 3.1.3 Hierarchical annotation

Using the manually-curated and cleaned dataset, we defined a hierarchy that is suitable for DL model training. It takes into account the availability of samples and the visual distinctness of classes. First, we defined the branches of the hierarchy, beginning with the root nodes *Insecta* and *Arachnida* and adding child nodes of descending taxonomic rank. The branches were extended to terminal nodes, the depth of which was dependent on the availability and granularity of the labelled data, while respecting the constraints of biological taxonomy. Three criteria governed the construction of the hierarchy: (i) taxonomic coherence, ensuring that all classes remain within their correct taxonomic branches; (ii) data availability, requiring a minimum number of labelled samples for a class to be viable for model training; and (iii) visual similarity, requiring that the members of a class share enough morphological features to be learnable by a computer vision model. Taxonomic coherence was treated as an inviolable constraint, while data availability and visual similarity jointly determined the depth to which each branch was resolved. This resulted in an asymmetric hierarchy where the depth of terminal nodes varies across branches: more abundant and visually distinguishable taxa (such as *Sepsidae* and *Vespidae*) were defined on branches to the taxonomic rank of family, while rarer or visually more homogeneous taxa (such as *Orthoptera* and *Mecoptera*) were defined only to the rank of order.

Second, nodes in the hierarchy tree were retained only for classes with uniquely labelled images at that node. This excluded redundant nodes, which would contain the same information as their parent or child nodes. An exception to this was for intermediate nodes with two or more closely-related child classes at the same level (for example *Coleoptera* has no unique images but is retained as a parent class of *Anthicidae, Chrysomeloidea, Coccinelloidea* and *Curculionoidea*). This resulted in a dataset with up to five levels of hierarchical labels per class. The coarsest level L1 comprises the classes *Insecta* and *Arachnida*. The respective levels of the hierarchy are not strictly associated with a taxonomic rank. The hierarchy is displayed in Figure 3.

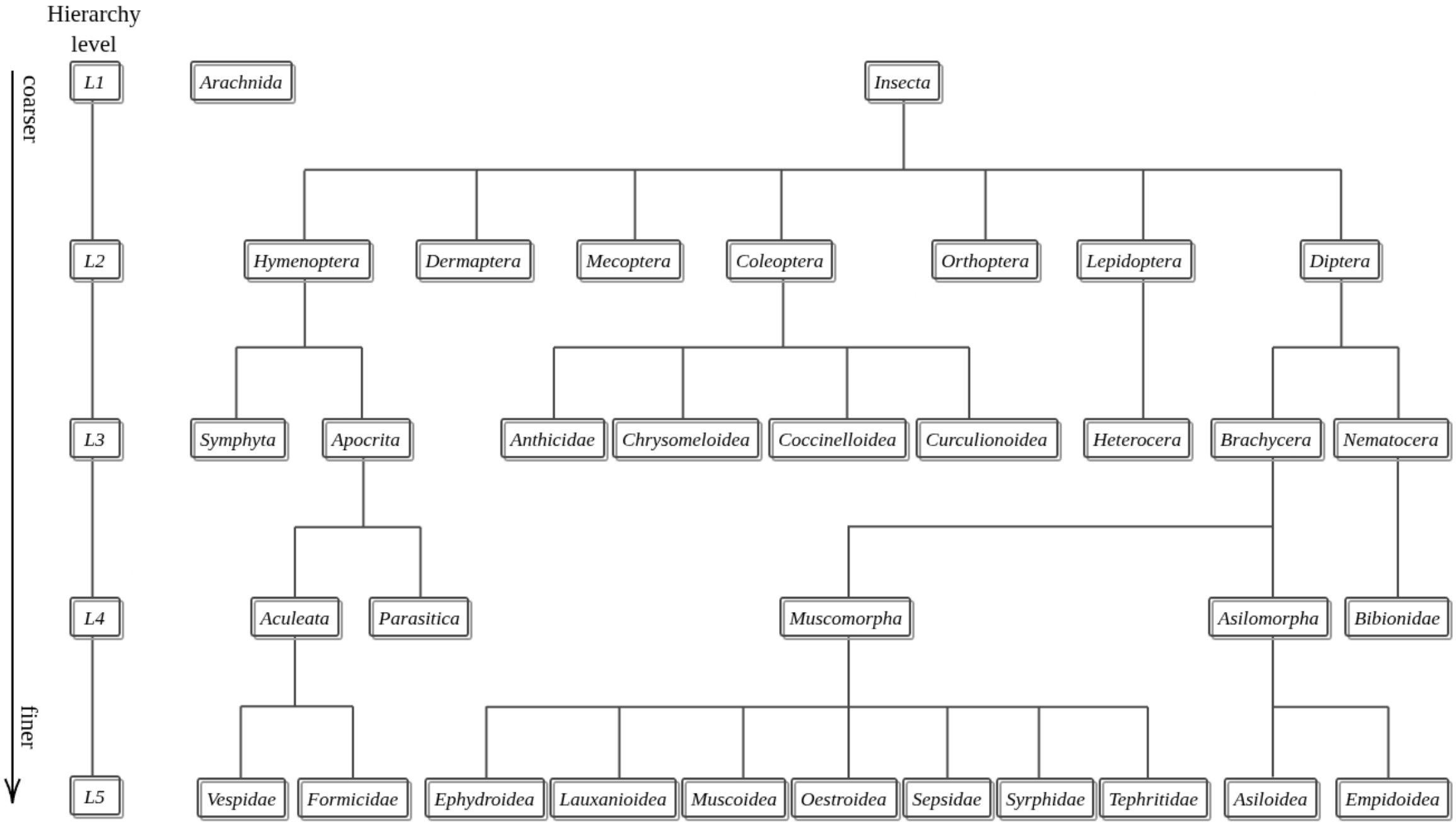

*Figure 3: Diagram of the hierarchy of the dataset, showing levels L1-L5, where L1 represents the coarsest level. Hierarchy levels are independent of taxonomic rank.*

The dataset is highly imbalanced (see Table S1 for full per-class video and image counts), with head classes of hundreds of videos and tail classes of as few as three videos. This is typical of real-world datasets where abundance varies greatly between classes and presents a particular challenge during training of the classification model. While it is beneficial to train on a large dataset to reduce the risk of over-fitting and improve the ability of the model to generalise to unseen samples, the imbalance can lead to bias in the model toward head classes. Corrections for this bias, which we account for in model development, are discussed in the following section.

## 3.2 Hierarchical insect classification model

The model architecture (visualised in Figure 4) is inspired by the method described by Chen *et al.* (2022), with several methodological changes. First, we expanded it to five classifier heads (one for each hierarchy level L1-L5), while adapting it to work with the variable-depth branches of our dataset. This ensures compatibility with samples with single labels (e.g. *Arachnida*, L1) and those with hierarchical labels (*Insecta* and its sub-classes, L1-L5). Additionally, we added both class-based and hierarchy-based weighting of the two loss functions (tree loss and cross-entropy loss), to account for the imbalance in the dataset.

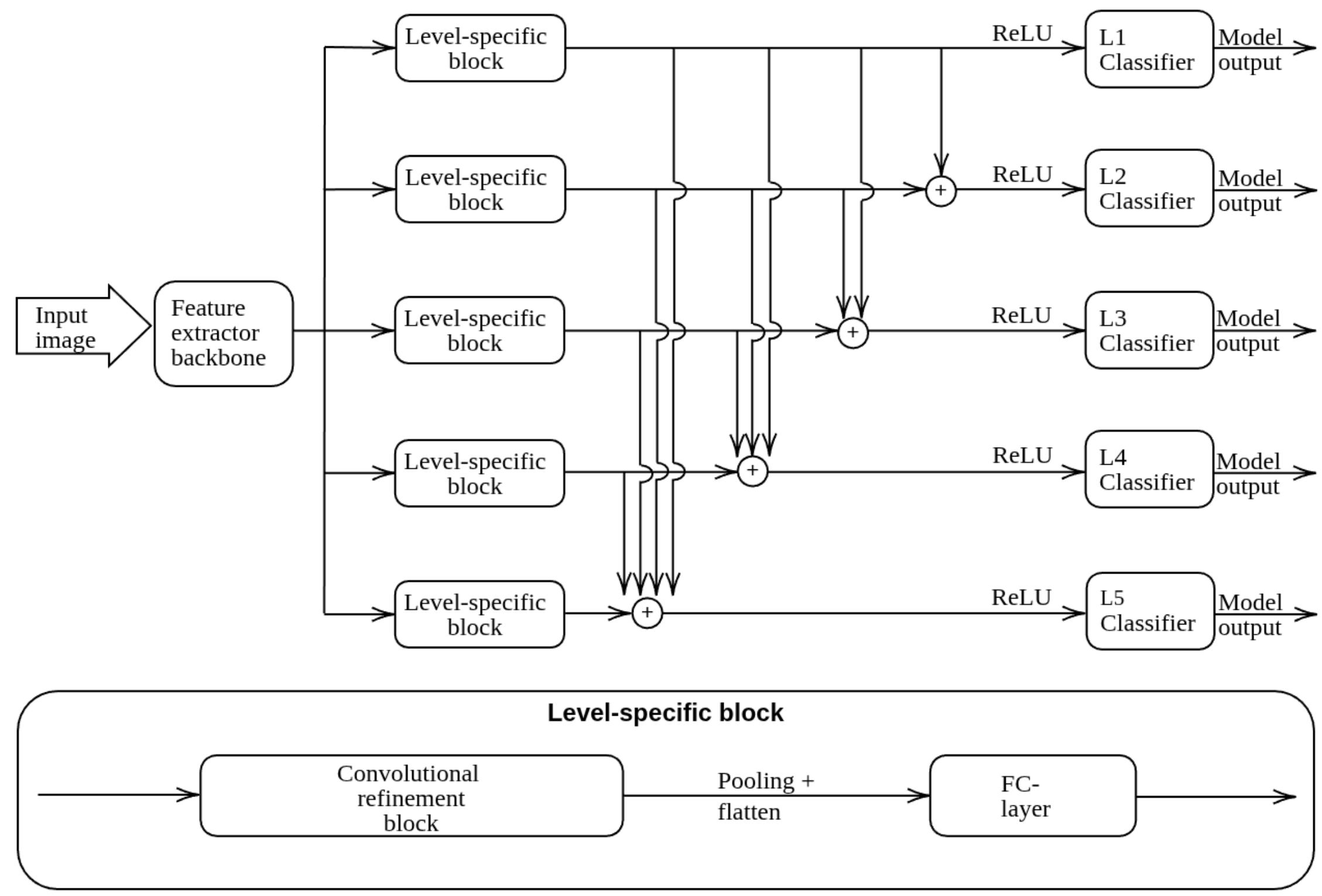

*Figure 4: Model architecture, incorporating a feature extractor, level-specific convolutional refinement blocks and fully-connected (FC) layers. FC-layer outputs are added to the outputs of finer levels of the hierarchy before ReLU activation and classification with a linear classifier.*

The model comprises a feature-extractor backbone, passing its output of global features to five level-specific blocks each comprising convolutional and fully connected layers which feed level-specific classifier heads. As a feature extractor we employed a pre-trained ResNet18, as a well-known and commonly used backbone, that works well on different tasks and is computationally lightweight. We performed end-to-end training, without freezing the backbone. Each level of the hierarchy has its own convolutional refinement block applied to the feature map output of the backbone. After pooling and flattening, each level in the hierarchy produces level-specific embeddings from its own fully-connected layer. The outputs of each level are added to the outputs of each of the finer levels below it, passing features from coarser to finer levels. ReLU activation is applied to each of the five (summed) outputs before classification with a linear layer at that respective level. Each classifier head predicts only among the classes defined at its own hierarchy level. The classifier outputs raw logits, to which sigmoid or softmax is applied for use in the tree loss and cross-entropy loss, respectively.

The loss function used to train the model comprises two additive components. The tree loss ensures hierarchical consistency and exploits the fine-grained features learned on sub-classes to influence predictions on their parent classes, while the cross-entropy loss improves separation between classes in the same hierarchical level. DL-based models require large amounts of data to be trained, both for adequate recognition of features and to avoid overfitting of the data. The latter arises where the model has the capacity to effectively 'memorise' all samples it has trained on and generalises poorly to unseen samples. Balancing the number of training samples to the least frequent class would dramatically reduce the size of the dataset available for training and increase the risk of overfitting. Instead, we applied the class-balanced weighting scheme described by Cui et al. (2019) to both loss functions to compensate for the imbalance in the number of samples. Rather than applying inverse class frequency directly, this approach increases the loss of classes with fewer samples to compensate for the dominance of head classes. Let $C$ denote the set of all hierarchy classes and $C_l \subseteq C$ the subset of classes belonging to hierarchy level $l$. For each hierarchy class $j \in C_l$ with $n_j$ training samples, an effective-number weight $\widetilde{w}$ is calculated as

$$\widetilde{w}_j = \frac{1-\beta}{1-\beta^{n_j}}, \tag{1}$$

where $\beta \in [0,1)$ controls the strength of class balancing.

These weights are then normalised to have a mean value of one:

$$w_j^{(node)} = \frac{\widetilde{w}_j}{\bar{w}}, \ \ where\ \bar{w} = \frac{1}{|C|} \sum_{k \in C} \widetilde{w}_k, \tag{2}$$

From preliminary experiments, we used an empirically determined value for $\beta$ (0.999995) to provide strong class balancing.

We first present the tree loss ($L_{hier}$; Equation 3), adapted from Chen et al., (2022). The tree loss employs a "state space" matrix

$$S \in R^{M \times |C|},$$

whose rows correspond to legal hierarchical states and whose columns correspond to classes in the hierarchy. Here $M$ denotes the total number of legal states represented in the hierarchy. For each sample $b$, the model produces a vector of sigmoid probabilities

$$F_b \in [0,1]^{|C|}.$$

For a ground-truth class $y_b$, let

$$A_{y_b} = \{ m \in \{1, \ldots, M\} \mid S_{m,y_b} > 0 \}$$

denote the set of legal states in which class $y_b$ is active. Each state $m$ is assigned a weight $w_m$, derived from the effective-number class weight $w_j^{(node)}$ defined in Equation 2. The partition function for sample $b$ is then

$$Z_b = \sum_{m=1}^{M} w_m e^{S_m^\top F_b}.$$

The hierarchical loss is then calculated as

$$L_{hier}(F, y) = \frac{1}{|B_{valid}|} \sum_{b \in B_{valid}} -\ln\left(\frac{\sum_{m \in A_{y_b}} w_m e^{(S_m^\top F_b)}}{Z_b}\right), \tag{3}$$

where

$$B_{valid} = \{b \in \{1, \dots, B\} \mid y_b \in C\}$$

is the set of samples with a valid class label for the hierarchy level under consideration.

The cross-entropy loss is applied independently to each level in the hierarchy and improves separation between classes on short branches and with fewer child classes, that benefit less from the hierarchical loss (e.g. *Dermaptera*, *Mecoptera*). The same effective-number weighting formulation (Equation 1) was applied to the cross-entropy loss, but normalised independently within each hierarchy level to maintain comparable loss magnitudes across levels. First, for hierarchy level $l$ we calculate the softmax probability $p$ of class $k$ for sample $b$:

$$p_{b,k}^{(l)} = \frac{e^{z_{b,k}^{(l)}}}{\sum_{j=1}^{|C_l|} e^{z_{b,j}^{(l)}}}, \tag{4}$$

where $z_{b,k}^{(l)}$ is the classifier logit for class $k$ at level $l$.

Mutual exclusivity of classes is particularly important at the coarser levels in the hierarchy, where a distinction between taxonomically distant orders (such as *Diptera* and *Hymenoptera*) is more critical to correct coarse-to-fine classification than between closely related genera. The cross-entropy loss was therefore weighted more strongly on the coarser level, using a variant of the method described in (Bertinetto et al., 2019) with initial $\alpha = 1.5$ decreasing in equal steps to 0.3 over 30 epochs. Let $d_l$ denote the depth of hierarchy level $l$, with the root level assigned depth 0. The level weighting $w_l$ is calculated with

$$w_l = e^{-\alpha \cdot d_l}, \tag{5}$$

then normalised across all levels $L$ of the hierarchy:

$$\widehat{W}_l = \frac{e^{-\alpha \cdot d_l}}{\sum_{r=1}^{L} e^{-\alpha \cdot d_r}}. \tag{6}$$

In addition to this per-level weighting $\widehat{W}$, we also apply the previously described per-class weighting $w$ (Equation Error: Reference source not found) to the cross-entropy loss and sum them to give the total cross-entropy loss $L_{CE}$ for all valid samples in the batch $B$:

$$L_{CE} = \frac{1}{|B_{valid}|} \sum_{b \in B_{valid}} \left( \sum_{l=1}^{L} \widehat{W}_l \cdot w^{(l)}_{y_b^{(l)}} \cdot \left( -\ln p^{(l)}_{b, y_b^{(l)}} \right) \right). \tag{7}$$

Finally, we sum the tree loss and cross-entropy loss to give the total loss:

$$L_{total} = L_{hier} + L_{CE}. \tag{8}$$

During inference, the same hierarchical state-space matrix $S$, introduced for the hierarchical loss during training, was used to score all legal taxonomic states. The sigmoid outputs from all hierarchy classes were concatenated into a single prediction vector $F_b$ for sample $b$. The score of legal state $m$ was then calculated as

$$q_{m,b} = S_m^{\top} F_b, \tag{9}$$

where $S_m$ denotes row $m$ of the state-space matrix. The most probable legal hierarchical state was then selected using maximum-a-posteriori (MAP) inference:

$$m_b^* = \arg\max_m q_{m,b} \tag{10}$$

The hierarchy classes active in state $m_b^*$ form a taxonomically consistent prediction path from the coarsest available level to the deepest predicted level. To accommodate varying levels of prediction certainty, a confidence-threshold strategy was applied. Starting from the deepest predicted class in the MAP path, the prediction was accepted if the sigmoid confidence associated with that class exceeded a predefined threshold, $T = 0.6$. Otherwise the prediction was truncated to the next coarsest level and the confidence was reassessed. We selected this threshold to maximise the number of confident predictions (see Section 4.2.2 for a comparison of probability thresholds). The threshold was not applied to L1, to ensure a prediction at the coarsest level was always made.

## 3.3 Experimental setup

In this section, we detail our experimental setup, including splitting the dataset for training validation and testing, the data augmentation methods that we applied, the metrics we used to assess the model's performance and the details of the hyperparameter optimisation.

### 3.3.1 Data splits and augmentation

When splitting the dataset into training, validation and test subsets, cropped images extracted from one video are assumed to be of the same specimen. Due to the motion-activated recording in the FAIR-Device, in several cases multiple videos of the same individual were made between entry into and exit from the device, in particular for specimens that moved intermittently. To avoid data leakage, a filter was applied to the dataset to ensure that images for a given class that were recorded on the same day and with the same device were only used in one of the subsets, as it was deemed that for such videos there would be a high likelihood that the same individual would be present in multiple videos. For this reason, the number of images (as opposed to the number of videos) was used as the number of samples, in particular for the calculation of class-based weighting in the loss function.

The number of samples per class and subset is detailed in Table S2. We applied a minimum of 450 images per class for training and validation, and 150 for testing, based on empirical checks that these provided sufficient samples. An upper limit of 18000 images for training and 6000 for testing, per class, was applied, employing random sampling while maintaining the within-day, within-device constraint described above to avoid data leakage. The upper limit was applied due to the extreme imbalance between head and tail classes (e.g. *Diptera* having 703 times as many images as *Dermaptera*). Nevertheless, the data splits remained highly imbalanced, as detailed in Table S2. The minimum and maximum numbers of images relate to the finest hierarchy level available for those samples. This means that the number of images for a class is the sum of the number of images for all its child classes plus the samples labelled only at its class. Due to the nature of the hierarchies with differing numbers of child classes for each parent class, imbalances exist in the number of images at coarser levels in the hierarchy, even if the child classes have a balanced number of images.

Transformation and augmentation of the samples was applied during training to augment the dataset and to improve the ability of the model to generalise. The transformations applied to the training set included resize, random crop, flip and rotation, colour jitter, random affine translation and *random_erase* (2-15%) before normalisation. Resize and centre crop were used on the test images before normalising.

### 3.3.2 Evaluation metrics

We report the coverage per level and per class, comprising the percentage of ground-truth samples for which predictions are made and which exceed the probability threshold. The performance of the model on the test set was assessed on a per-class and per-level-of-hierarchy basis. Per-class metrics of precision, recall, F1 and average confidence were recorded, to determine the model's ability to distinguish between often visually similar classes. We also tested the hypothesis that incorrect predictions in a hierarchical model are more likely to be taxonomically related to the ground-truth class than to an unrelated class. Here we reported what percentage of incorrect predictions (false negatives of a class) share a common parent class with the correct class. For example, if a sample with a ground-truth label *Oestroidea* is misclassified, we report what percentage of those incorrect predictions are members of *Muscomorpha* (the parent class of *Oestroidea)*.

On a per-level basis, performance is measured with accuracy-on-covered[2] and Matthews correlation coefficient (MCC). While a model that defaults to predicting the dominant class all the time may have high accuracy, a high MCC is dependent on all classes, including minority classes, being correctly predicted.

### 3.3.3 Hyperparameter optimisation and model training

Training over 100 epochs followed a progressive resize schedule for the input images of 96x96 px (epochs 0-9), 112x112 px (epochs 10-19) and 128x128 px (epochs 20-99). We based the starting input size on the modal image size of the dataset (90x90 px for 2023 data, see Figure S1). Optimisation used Stochastic Gradient Descent with momentum of 0.9 and weight decay 0.001. A cyclical learning-rate scheduler with a peak learning rate of 2.5 times the initial rate and 40 epochs warm-up was used. Regularisation with dropout of 0.2 and 0.5 was used on the convolutional and fully-connected layers, respectively. Optimisation of the backbone learning rate (range: 0.00005-0.01), classifier learning rate (range: 0.00001-0.01) and batch size (range: 512-4096) was carried out using evolutionary multi-objective optimisation on the training and validation splits. This method uses search algorithms to simultaneously optimise multiple objectives, maintaining a diverse set of solutions representing trade-offs, known as the Pareto front. New trials dominate existing ones when they are at least as good on all objectives and better on at least one objective. The objective was to maximise the MCC on the validation set, for each of the five levels in the hierarchy. We ran an initial experiment of 30 trials, which showed a pattern of convergence and added additional trials based on available compute resources (total *n_trials* = 77). The model was retrained with the combined training and validation data and tested on the unseen test set with the parameters from the optimisation trial with highest averaged validation MCC (backbone learning rate: 0.00043, classifier learning rate: 0.00075, batch size: 1024). Code was written in Python and the model was trained on a server featuring an AMD EPYC 9334 32-Core Processor with 1.1TiB of RAM, equipped with four NVIDIA RTX 4500 Ada Generation GPUs

[2] Accuracy-on-covered denotes the accuracy of predictions, based on the subset of samples for which a prediction exceeds the defined threshold of confidence (i.e., the coverage).

# 4 Results and discussion

In this section we first provide an overview of the results of our hierarchical invertebrate model (see Section 3.2), developed and tested on the manually-curated dataset we present (see Section 3.1.2). We then provide a detailed analysis of per-class and per-level performance, including rare-class classification, visually similar taxa, and the effect of the confidence threshold, before discussing our results in the context of related work and outlining limitations and future directions.

## 4.1 Results overview

The performance of our arthropod classification model is summarised in Table 1, showing the per-level and per-class metrics for the test set. The model achieved a minimum per-level accuracy-on-covered of 0.8174 (range 0.8174 – 0.9882). The MCC scores indicate that the evenness of predictions improved from L1 through L5. We would expect the MCC to be higher at L5, the most balanced level, but even for the highly imbalanced L2 level the MCC of 0.7198 demonstrates that the class-balancing weighting seems to be effective. The appropriateness of MCC as a key metric is confirmed at the coarsest level of the hierarchy (L1), where the combined accuracy of over 98 percent hides the fact that the rare class *Arachnida* has a recall score of 0.1682 and so is often misclassified as *Insecta*. The low recall for *Arachnida* is likely a combination of the effect of low sample numbers ($n$ = 829) and significant occlusions on images for this class. These are discussed in more detail in Section 4.2.3.

*Table 1: Per-level and per-class metrics for the test set, where L1 represents the coarsest level of root classes (Insecta and Arachnida) and L5 the finest level. Coverage, Accuracy-on-covered and MCC are reported per level. Coverage, Precision, Recall and F1e are reported per class. Additionally, for levels L3-L5 we provide a metric on the frequency that false negatives for a class share a common parent with the ground truth label. Confidence threshold T = 0.6.*

| Hierarchy level | Coverage (%) ↑ | Accuracy on covered ↑ | MCC ↑ | Class | Class Coverage (%) ↑ | Precision ↑ | Recall ↑ | F1 ↑ | False negative with common parent class ↑ |
|---|---|---|---|---|---|---|---|---|---|
| L1 | 100 | 0.9882 | 0.2489 | *Insecta* | 100 | 0.9909 | 0.9973 | 0.9941 | - |
| | | | | *Arachnida* | 100 | 0.3959 | 0.1628 | 0.2308 | - |
| L2 | 93.80 | 0.8317 | 0.7198 | *Coleoptera* | 88.22 | 0.8246 | 0.6197 | 0.7076 | - |
| | | | | *Diptera* | 98.05 | 0.8373 | 0.9087 | 0.8715 | - |
| | | | | *Hymenoptera* | 96.21 | 0.8185 | 0.6525 | 0.7261 | - |
| | | | | *Lepidoptera* | 83.68 | 0.6482 | 0.4036 | 0.4974 | - |
| | | | | *Dermaptera* | 16.14 | 0.4894 | 0.1614 | 0.2427 | - |
| | | | | *Orthoptera* | 82.42 | 0.9750 | 0.7756 | 0.8639 | - |
| | | | | *Mecoptera* | 78.19 | 0.6579 | 0.6597 | 0.6588 | - |
| L3 | 95.54 | 0.8174 | 0.6840 | *Apocrita* | 98.19 | 0.8141 | 0.7838 | 0.7986 | 0.0053 |
| | | | | *Brachycera* | 99.06 | 0.8285 | 0.9198 | 0.8717 | 0.0687 |
| | | | | *Nematocera* | 90.19 | 0.8425 | 0.6455 | 0.7310 | 0.3425 |
| | | | | *Anthicidae* | 85.39 | 0.5911 | 0.6742 | 0.6299 | 0.0 |
| | | | | *Coccinelloidea* | 94.92 | 0.9527 | 0.8493 | 0.8981 | 0.1536 |
| | | | | *Chrysomeloidea* | 73.50 | 0.0 | 0.0 | 0.0 | 0.0349 |
| | | | | *Heterocera* | 84.73 | 0.6201 | 0.3792 | 0.4706 | 0.0 |
| | | | | *Symphyta* | 87.37 | 0.9052 | 0.1553 | 0.2651 | 0.092 |
| | | | | *Curculionoidea* | 73.42 | 0.4456 | 0.1827 | 0.2556 | 0.0108 |
| L4 | 95.82 | 0.8728 | 0.7814 | *Aculeata* | 96.38 | 0.8930 | 0.7609 | 0.8217 | 0.3434 |
| | | | | *Asilomorpha* | 99.47 | 0.8877 | 0.8060 | 0.8449 | 0.8473 |
| | | | | *Muscomorpha* | 95.68 | 0.9045 | 0.8876 | 0.8960 | 0.2714 |
| | | | | *Bibionidae* | 99.82 | 0.8224 | 0.9584 | 0.8852 | 0.0 |
| | | | | *Parasitica* | 84.28 | 0.5112 | 0.5369 | 0.5237 | 0.1563 |
| L5 | 92.54 | 0.8230 | 0.7966 | *Formicidae* | 89.50 | 0.9852 | 0.7670 | 0.8625 | 0.0 |
| | | | | *Empidoidea* | 75.32 | 0.3684 | 0.03636 | 0.0662 | 0.3080 |
| | | | | *Oestroidea* | 96.10 | 0.8224 | 0.7462 | 0.7824 | 0.7493 |
| | | | | *Ephydroidea* | 99.59 | 0.8907 | 0.3869 | 0.5394 | 1.0 |
| | | | | *Syrphidae* | 91.05 | 0.8626 | 0.7547 | 0.8050 | 0.5070 |
| | | | | *Lauxanioidea* | 82.63 | 0.9652 | 0.7339 | 0.8338 | 0.9942 |
| | | | | *Muscoidea* | 97.12 | 0.6313 | 0.9083 | 0.7449 | 0.7342 |
| | | | | *Asiloidea* | 98.62 | 0.8945 | 0.8412 | 0.8670 | 0.0 |
| | | | | *Sepsidae* | 95.10 | 0.9559 | 0.1275 | 0.2249 | 0.9929 |
| | | | | *Tephritidae* | 98.25 | 0.7036 | 0.5664 | 0.6276 | 1.0 |
| | | | | *Vespidae* | 87.23 | 0.6277 | 0.7220 | 0.6715 | 0.0196 |

## 4.2 Results Analysis

Following the overview of results presented in Section 4.1 we will now present a more detailed analysis of our findings, including the effects of class balancing to correct for dataset imbalance, the performance of the model on visually-similar insects and the impact of the probability threshold on the number of predictions per level.

Looking at the results for the individual classes, 13 of the 34 classes in the test set have F1-scores above 0.8. High F1-scores are observed at all hierarchical levels and irrespective of branch depth. To assess the overall effect of class-balanced weighting, we plotted the number of sample images per class against the test F1-scores (see Figure 5). The number of images per class explains less than one third of the variation of test F1-scores ($R^2$ = 0.3520), suggesting that a model can perform well on the tail classes of a highly-imbalanced dataset, when class-balanced weighting is applied during training. For example, for the class *Bibionidae* (*n* = 3,060), representing only 1.15 percent of the dataset, the model achieved an F1-score of 0.8852. Furthermore, the 3,060 images of *Bibionidae* were extracted from four videos of one individual recorded over ten minutes on 24.08.2023, while the test images were from one video of one individual, comprising 1 minute 35 seconds of video recorded on 30.08.2023. This indicates that the model is capable of highly accurate prediction on individual classes even when trained on few samples. We note, however, that this result reflects a single individual and may not generalise to specimens photographed at different sites, seasons, or under different lighting.

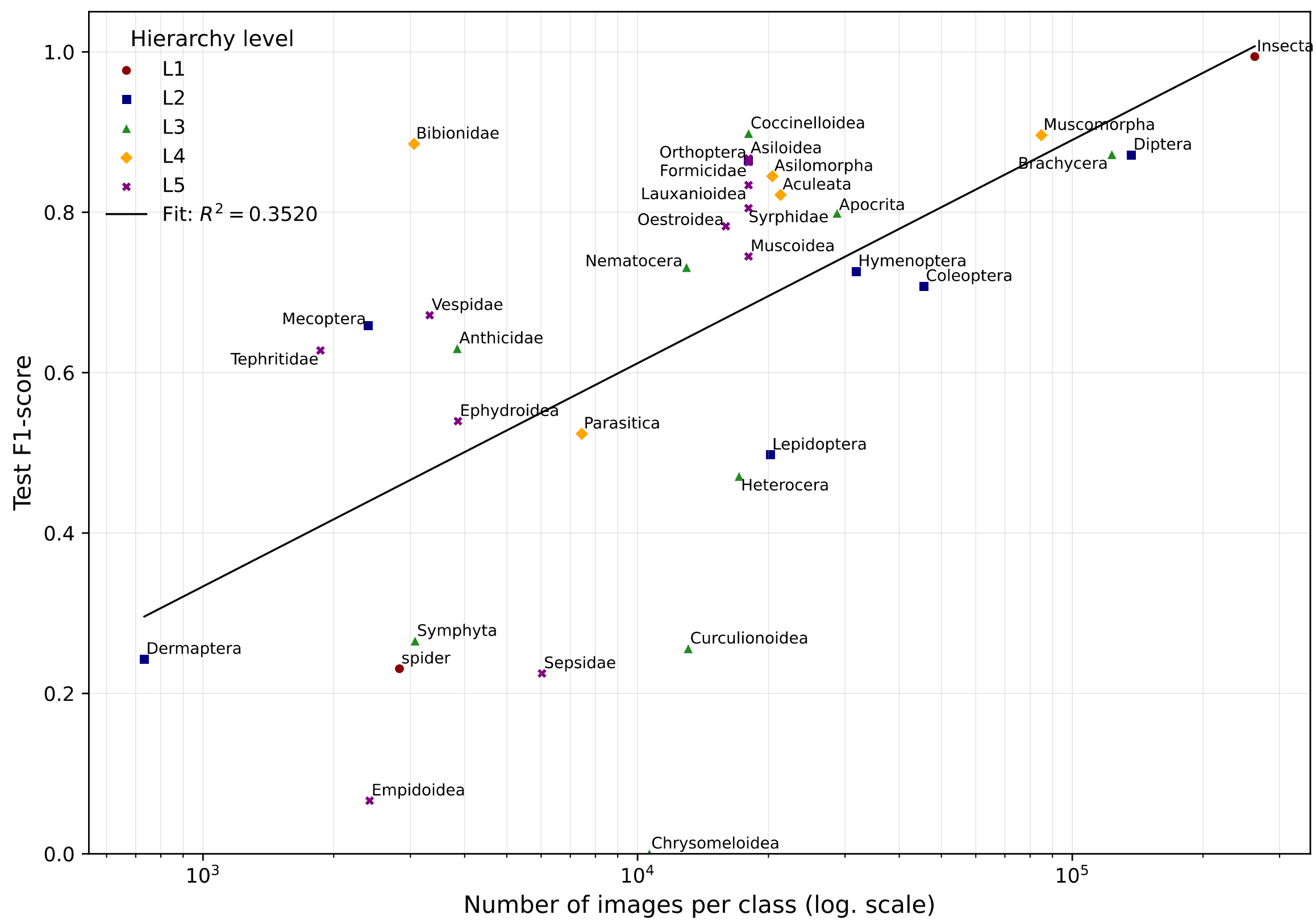

*Figure 5: F1-score as a function of the number of images per class (logarithmic). The number of images per class correlates weakly with the F1-score. The model also achieves high F1-scores on rarer classes, such Bibionidae, indicating that number of images is not the sole determinant of per-class performance.*

### 4.2.1 Distinguishing between visually similar insects

Since the classification model we developed solely relies on image data, visual similarity of different insects is a potential reason for misclassifications. Considering the visually-similar but distantly-related *Vespidae* and *Syrphidae*, the model misclassifies 334 of 6000 samples (5.57 percent) of *Syrphidae* as *Vespidae* and 92 of 1018 samples (9.04 percent) of *Vespidae* as *Syrphidae*. Individually, the classes have recall scores of 0.7547 for *Syrphidae* and 0.7220 for *Vespidae*.

For the closely related and often visually similar subclasses of *Muscomorpha* (*Oestroidea*, *Ephydroidea*, *Tephritidae*, *Lauxanioidea*, *Sepsidae*, *Syrphidae* and *Muscoidea*) our model achieved high precision scores (range: 0.6313-0.9652). When misclassification does occur, there is a high tendency for these classes to be misclassified as *Muscoidea*, rather than a more distantly-related class (see the L5 confusion matrix in Figure 6). This is confirmed by the high scores for these classes for false negatives with a common parent class in Table 1.

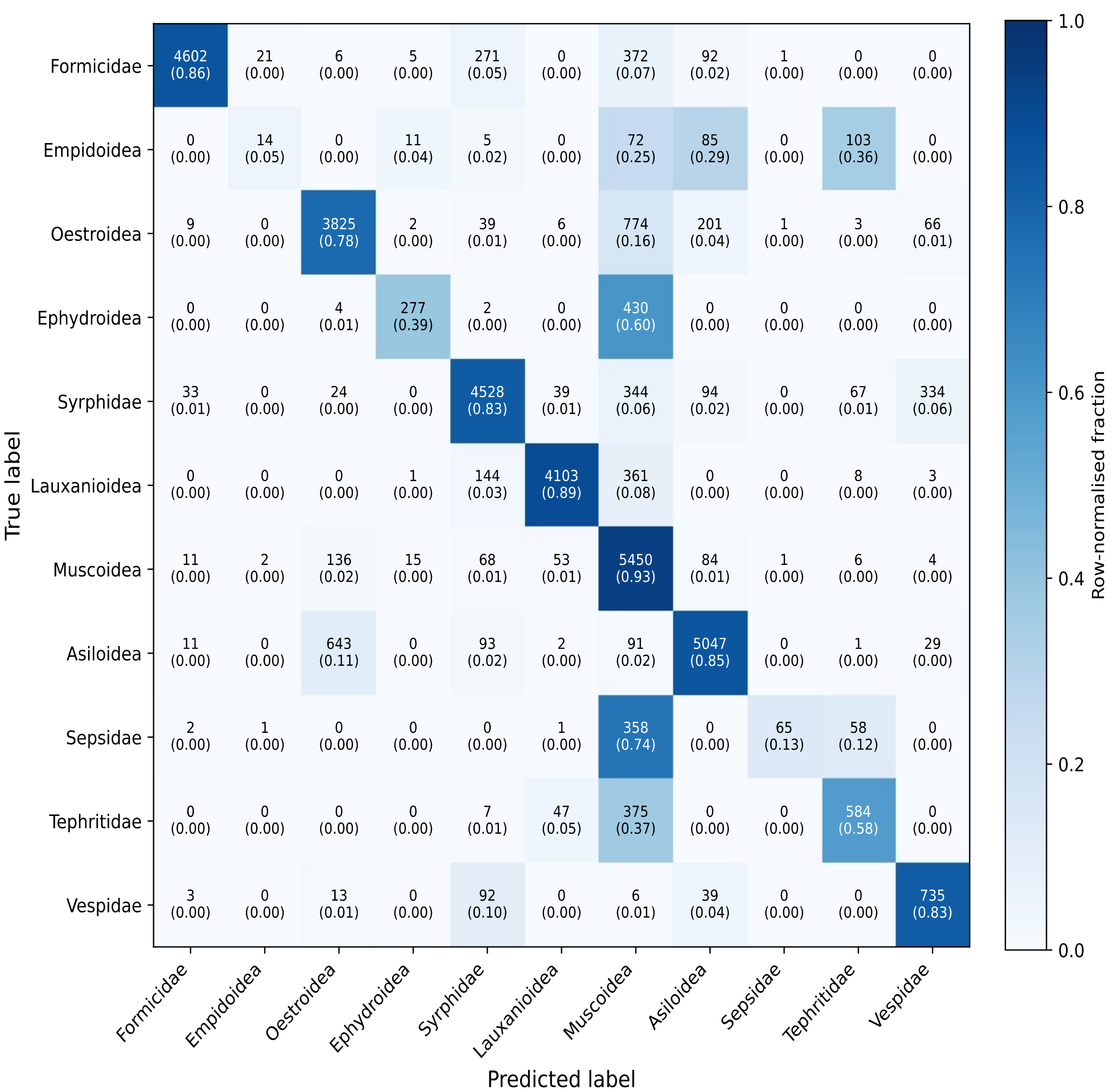

*Figure 6: Confusion matrix of L5 of the hierarchy. Rows represent the ground truth, columns represent predictions exceeding the confidence threshold (T = 0.6). Cells show sample counts and, in parentheses, row-normalised fractions. Colours indicate row-normalised fractions.*

### 4.2.2 Classification at different hierarchy levels

The accuracy of the model to predict to the correct depth in the hierarchy, as well as the effect of the threshold of probability on the number of predictions per level is detailed in Table 2. The model predicts to a shorter depth than the ground truth label in seven percent of samples at the deepest level of the hierarchy (L5), decreasing to 0.28 percent for the L2 level. Predictions for levels deeper than their assigned label were disregarded as they had no ground truth to validate the correctness of predictions. For predictions with valid ground truth labels, a threshold of $T = 0.6$ resulted in at least 92.54 percent of predictions being made without reverting to a coarser level. At the L2 level, almost six percent of predictions do not meet the threshold; without the hierarchical fallback mechanism, these samples would have no valid prediction at all.

*Table 2: Percentage of predictions made at each hierarchy level that meet or exceed the defined probability threshold (T = 0.6)*

| Hierarchy level | Number of samples | Prediction depth before thresholding | | | Samples below threshold ($T$ = 0.6) (%) | Coverage (predictions exceeding threshold ($T$ = 0.6) (%) |
|---|---|---|---|---|---|---|
| | | Shorter than ground truth (%) | Same as ground truth (%) | Longer than ground truth (%) | | |
| L1 | 76,384 | 0.00 | 96.80 | 3.20 | 0.00 | 100 |
| L2 | 73,805 | 0.28 | 96.55 | 3.16 | 5.92 | 93.80 |
| L3 | 66,046 | 0.32 | 78.13 | 21.50 | 4.13 | 95.54 |
| L4 | 42,676 | 3.29 | 92.90 | 3.80 | 0.89 | 95.82 |
| L5 | 38,377 | 7.00 | 93.00 | 0.00 | 0.46 | 92.54 |

We additionally compared the effect of varying threshold levels on the coverage and accuracy-on-covered (illustrated in Figure 7). Aside from level L1, where thresholding is not applied, an increased probability threshold results in reduced coverage and increased accuracy-on-covered. Coverage for levels L2-L5 is highest for $T$ = 0.4 (range: 0.93-0.98), while being substantially lower when tested at the higher probability threshold of $T$ = 0.9 (range: 0.77-0.89). As the probability threshold increases, the accuracy of predictions meeting this threshold increases. Accuracy-on-covered for levels L2-L5 is highest for $T$ = 0.9 (range 0.85-0.92), while for $T$ = 0.4 accuracy is in the lower range of 0.80-0.87.

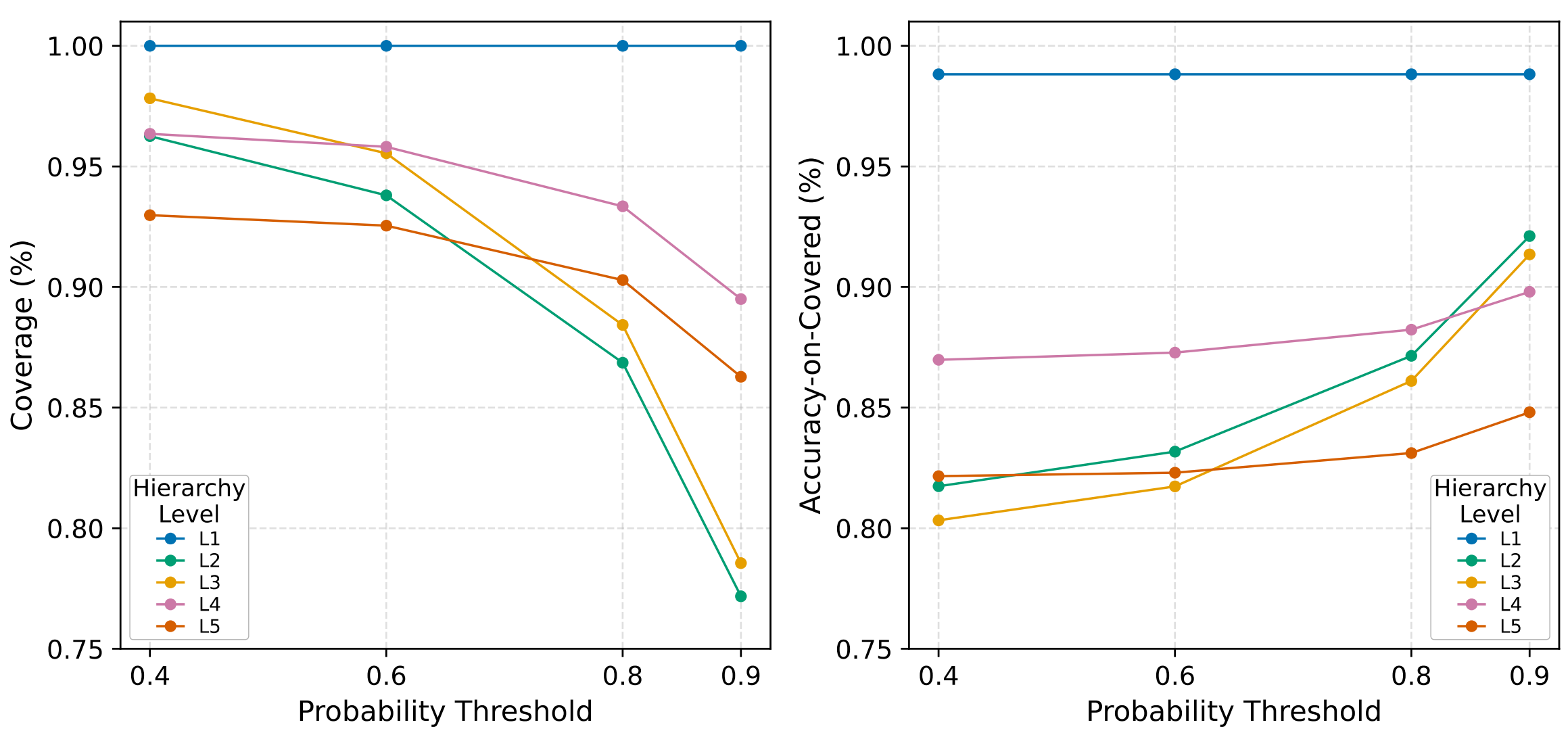

*Figure 7: Coverage (left panel) and accuracy-on-covered (right panel) for a range of probability thresholds (T = 0.4, 0.6, 0.8, 0.9) for each level of the hierarchy (L1-L5). Coverage decreases, while accuracy increases, with increasing threshold values.*

### 4.2.3 Model Limitations

Having presented the results and analysis of the model we now analyse some of the limitations of the model, in particular for classes that had particularly low F1-scores and possible reasons for this.

While balancing the loss function to improve learning on rare classes can be effective for highly imbalanced datasets, a lack of data still negatively affects the model's ability to correctly predict rare classes. This can be seen in Figure 8, where the number of images per class multiplied by the average image size i.e., the total number of pixels seen by the model per class, is plotted against the F1-scores. The average image size is capped to the maximum input size of the model, not the raw image size. The total pixels per class account for more of the variation ($R^2$ = 0.4466) of the F1-score than number of images per class alone and suggests that the model does not learn well on classes with low sample counts combined with low resolution.

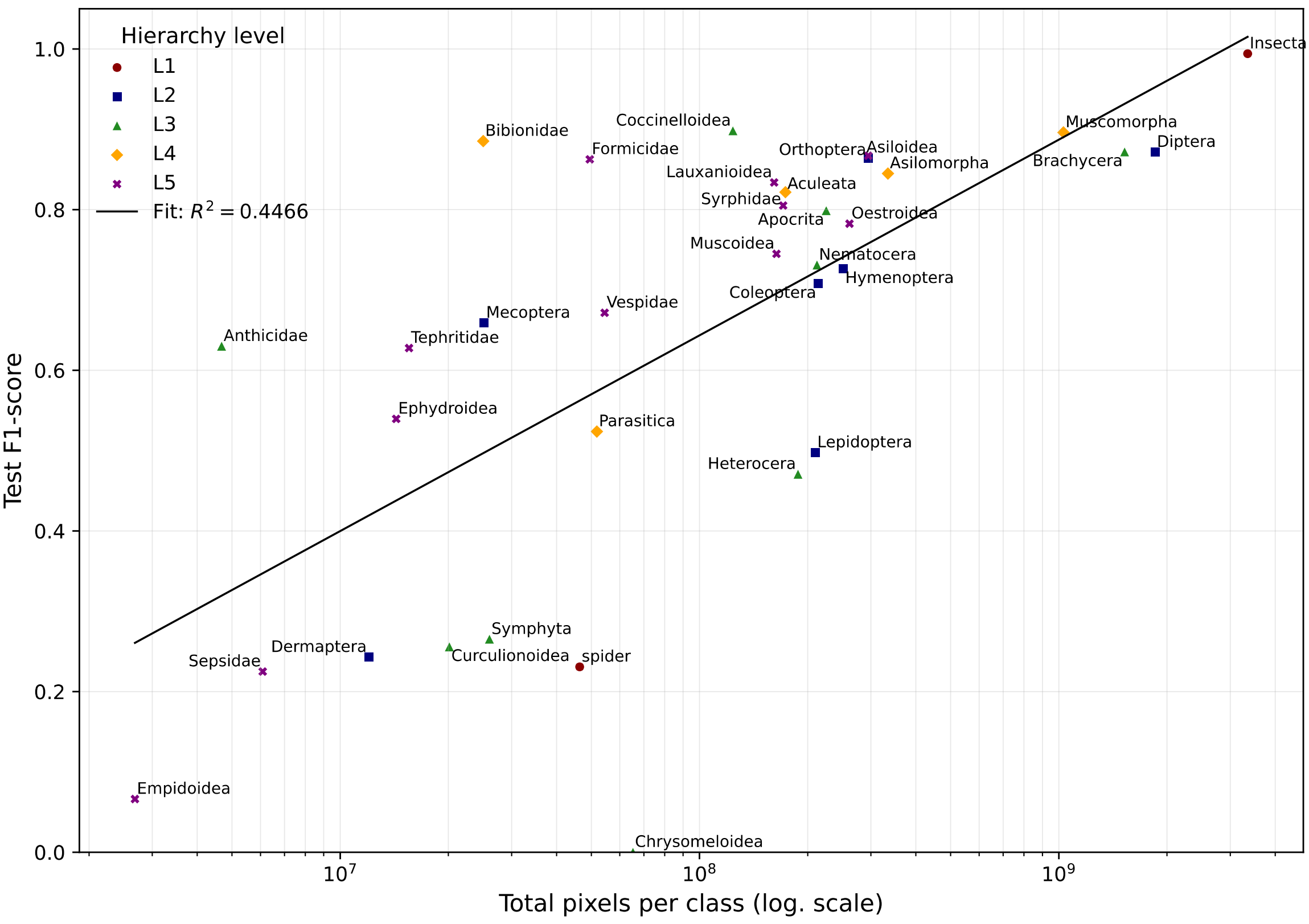

*Figure 8: F1-score as a function of the total number of pixels per class (number of images*image size, logarithmic scale). Total pixels per class explains more of the variation in F1-scores than number of images per class.*

Extensive augmentations and transformations were applied during training to reduce the risk of overfitting and improve the ability of the model to generalise to unseen data. One augmentation technique we employed is *random erasing*, which randomly replaced two to 15 percent of the sample image with a light grey rectangle, thus forcing the model to learn from other visible features. While generally a beneficial method, Figure 3 shows an example of specimens of *Dermaptera* and *Arachnida*, which already suffer from occlusion from the camera trap itself. In these cases *random erasing* may remove too much detail for effective learning of features.

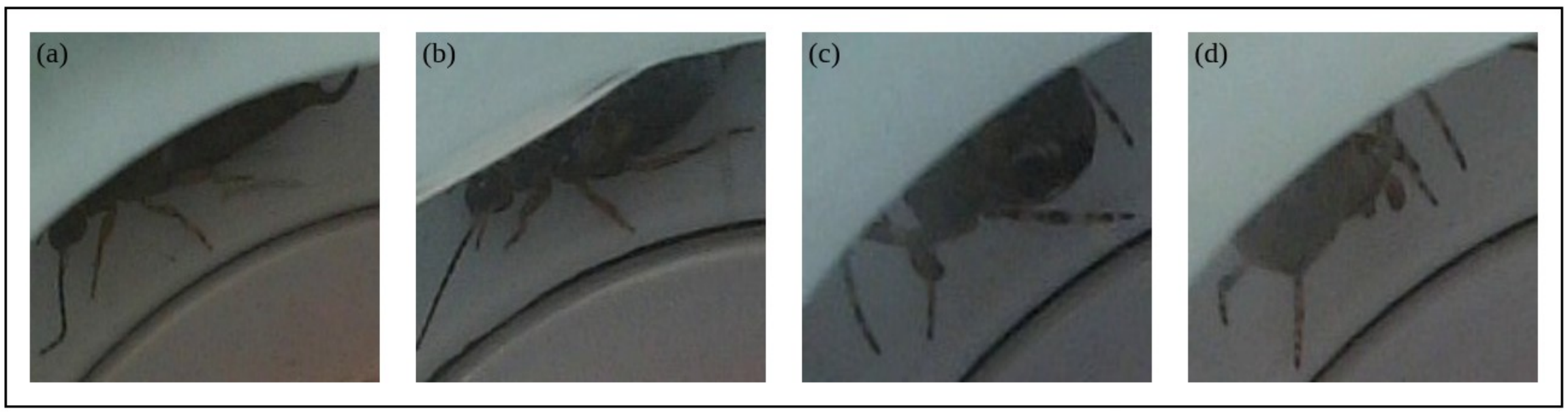

*Figure 9: Occluded specimens of the classes Dermaptera (a) and (b) and Arachnida (c) and (d).*

The leaf nodes of the hierarchy represent taxa spanning multiple species, genera or families, that share a common parent node (see Section 3.1.3). Although members of a class are likely to be visually similar, the limitations of this approach must be recognised when working with classes of high intra-group variation and low sample numbers. This is illustrated in the case of the class *Chrysomeloidea* (Figure 10) that had precision, recall and F1-scores during testing of 0.0 (*n_samples* = 351). The four specimens labelled as *Chrysomeloidea* are visually different to one another, to the extent that the model did not learn common identifying features and misclassified all samples in the test set.

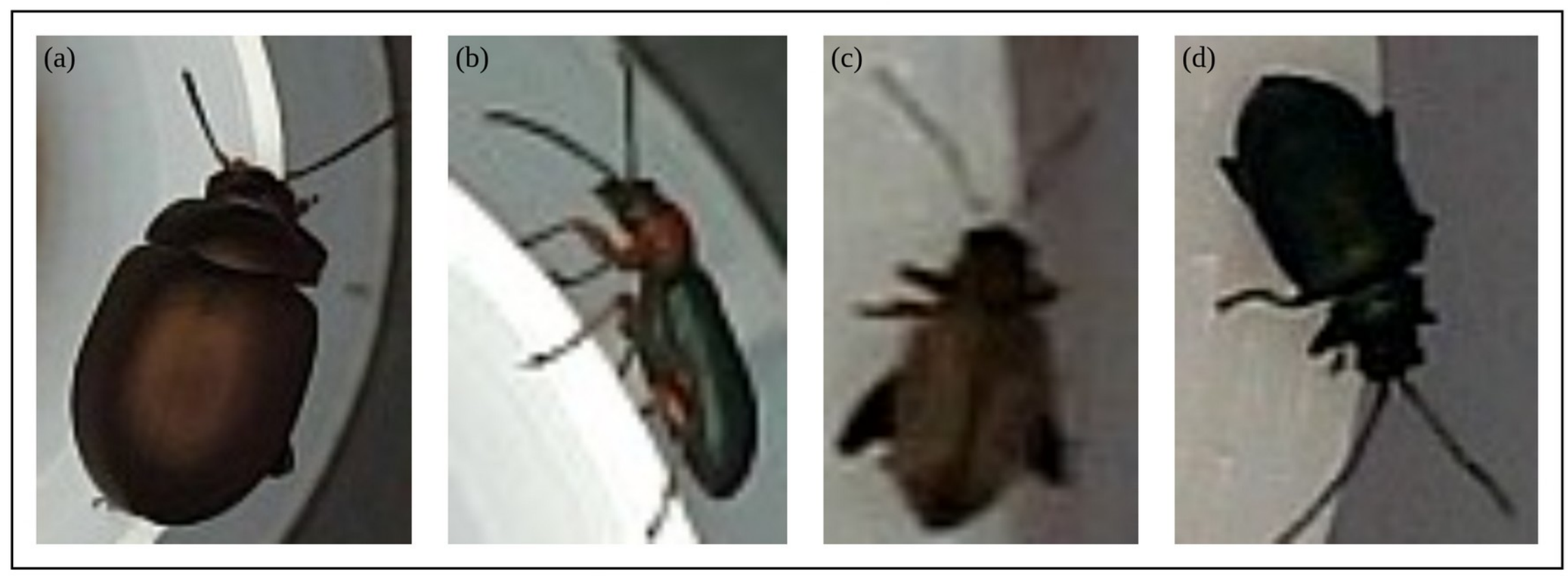

*Figure 10: The four specimens of Chrysomeloidea demonstrate the domain shift between train and test splits. Specimens (a) and (b) were used in training, samples (c) and (d) in testing.*

## 4.3 Discussion

The potential for automated insect camera traps to generate data (e.g. video, time-lapse photographs, cropped images) for model training has been used by other researchers (Bjerge et al., 2023a; Sittinger et al., 2024) and in our case resulted in a curated dataset of almost one million images. In contrast to the downward-facing cameras observing a horizontal platform or substrate, used in the two referenced works (see Figure 1, (a) and (b)), our dataset is based on videos of insects passing through a camera trap. This resulted in highly varied poses of the specimens, which benefits training data. With accurate tracking of specimens, a single labelled video can produce thousands of cropped images for model development. While this can alleviate a key bottleneck to DL-based model development – namely a lack of sufficient labelled data – accurately labelling specimens in videos or photographs still requires expert curation to ensure reliability of the data. This in particular is the case for very small insects, whose identification for labelling may be limited by the resolution of the images.

Our results demonstrate the architecture's ability to classify rare classes and visually similar classes, as well as correcting for dataset imbalance. This is notable relative to other hierarchical and flat models, where training on highly imbalanced datasets has not been directly addressed. Despite these positive results, the model is limited by the data available for training. Two situations resulted in poor performance: (i) classes with low resolution combined with low sample numbers and (ii) low-sample classes where all specimens are visually distinct from one another. Changes to the model architecture and increasing the camera resolution may address the first situation. Increased resolution would, however, require more storage capacity for recorded images, increased processing demands for on-device classification and may shift the limit of resolution to even smaller insects that heretofore have not been part of the curated dataset. Depending on the use case, the exact determination of insects will always be limited with camera traps, as distinguishing features may not be visible to the device. Regarding the issue of visually-heterogenous classes, careful consideration must be given during annotation to taxa with high intra-class variability, increasing the sample size or merging these taxa with their parent class.

Our adaptation of the model architecture from Chen et al. (2022) to a 5-level hierarchy confirms the flexibility of this architecture. We observed strong performing classes at every level, or depth, of the hierarchy. While other examples of hierarchical classification assigned a specific taxonomic rank to each level of the hierarchy (e.g. order-family-species) (Bjerge et al., 2023b; Chang et al., 2021; Chen et al., 2022; Elhamod et al., 2022), our model implements a more flexible approach that can be adapted to available labels and does not require all classes to be labelled to species level. The progressive reduction in the number of unclassified samples, through the application of the probability threshold and reclassification at coarser levels, underscores the usefulness of multi-granularity classification to minimise uncertain predictions and unclassified samples. Additionally the choice of threshold allows for the number of, or confidence of, predictions to be prioritised. However, for all the benefits of a hierarchical model, expanding the hierarchy to include more levels must be assessed critically in the context of the available data. Each additional level adds complexity to the model, increasing the number of trainable parameters and thus the need for more data and/or augmentations and adjustment of training schedules to avoid overfitting.

While this work has created a new hierarchical dataset of insects and performed well on the test subset of this dataset, future work could test the model on unseen species from the same parent groups as this dataset. The results presented here are also based on a limited number of taxa collected in Germany, although they represent many of the common orders of insects found worldwide.

Monitoring of insects with automated camera traps requires a number of distinct processes to work together, including specimen detection, tracking, classification and enumeration. The classification method presented in this paper needs integration with the other processes to fully realise its potential in real-time biodiversity monitoring. This would require further research to ensure, not only that these task-specific components are compatible with one another, but also that they fulfil the requirements for working in the field. While this model uses a relatively lightweight ResNet18 backbone, comparison with other backbones would facilitate the decision between model accuracy and suitability for running in low-powered, remote monitoring devices. To this end the model has already been coded to be compatible with a number of backbones, including other ResNet (He et al., 2015) and MobileNet (Howard et al., 2017) feature extractors. Beyond this, pruning and quantizing the model to further reduce power consumption and optimise it for running on small edge devices could be undertaken to facilitate its use in remote automated insect camera traps.

# 5 Conclusion

Applying DL to insect classification for biodiversity monitoring requires expert-annotated data that are costly and time-consuming to acquire. At the same time, flat classification models are known to struggle to generalise well for large numbers of classes of varying granularity and imbalanced training data. We address these challenges with a hierarchical dataset extracted from automated camera trap recordings and a hierarchical image classification model with class-balanced weighting. Our model performs well on head and tail classes and distinguishes between visually similar classes. Furthermore we leverage the hierarchy to maximise the confidence of predictions and minimise unclassified samples. We hope that our work, i.e., both the dataset we constructed and the model we developed, will serve as the basis of future research in the context of AI-based biodiversity monitoring.

# 6 Declaration of generative AI and AI-assisted technologies in the manuscript preparation process

During the preparation of this work, the authors used Microsoft Copilot to assist with improving the clarity, consistency and completeness of the mathematical notation and descriptions of the hierarchical loss and inference procedures described in Section 3.3. After using this tool, the authors reviewed and edited the content as needed and take full responsibility for the content of the published article.

### CRediT authorship contribution statement

Zaki Mahfoud: Methodology; Software; Validation; Data curation; Formal analysis; Writing – original draft; Visualization

Juan A. Chiavassa: Conceptualization; Methodology; Validation; Data curation; Writing – review & editing;

Simon Walther: Writing – review & editing; Supervision; Project administration; Funding acquisition.

Florian Haselbeck: Conceptualization; Methodology; Writing – review & editing; Supervision; Project administration;

Ehsan Yaghoubi: Conceptualization; Methodology; Validation; Writing – review & editing; Supervision;

### Funding

We thank the Free State of Bavaria for funding the positions affiliated with the Professorship of Smart Farming within the framework of the Hightech Agenda Bavaria.

The camera-trap device used to acquire the raw dataset used in this paper was developed as part of the joint project “Monitoring of biodiversity in agricultural landscapes” (MonViA) that was funded by the German Federal Ministry of Food and Agriculture (BMEL).

### Declaration of competing interest

The authors declare that they have no known competing financial interests or personal relationships that could have appeared to influence the work reported in this paper.

### Data and code availability

Dataset: available at: 10.5281/zenodo.21690971
Python Code: https://github.com/smAIL-WS/HierarchicalInsectClassification

## Supplementary material

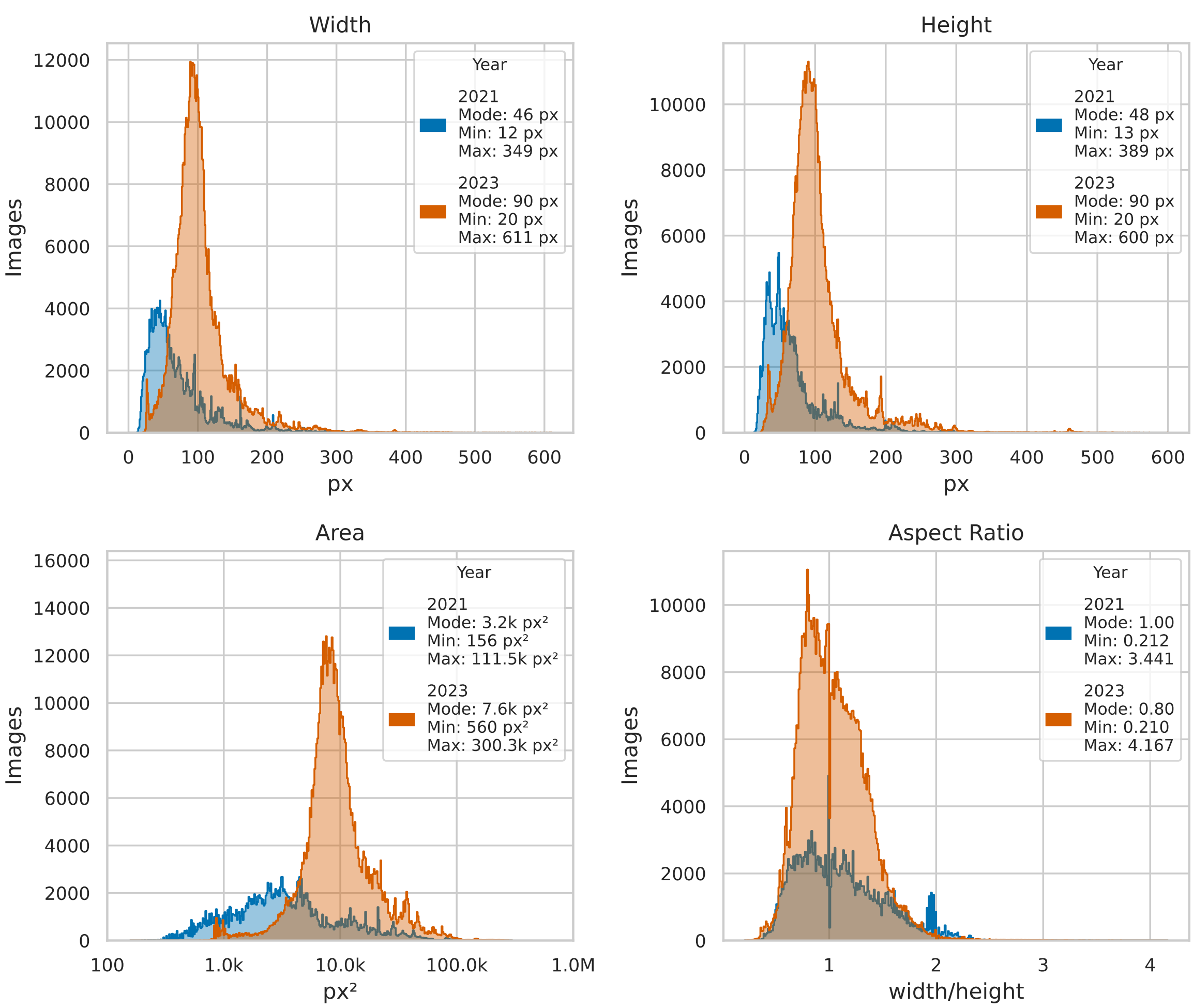

*Figure S1: Graphs of cropped image size and aspect ratio, by year. A higher-resolution camera used in 2023 resulted in larger image sizes. The dataset exhibits large variation in image size (range: 156 px$^2$ - 300,300 px$^2$) and aspect ratio (range: 0.21 - 4.167), requiring transformation to avoid distortion during resizing to the model input size.*

*Table S1: Number of source videos and cropped images per class, sorted by hierarchy level. Highest and lowest numbers of source videos and cropped images for each level are marked green and red, respectively.*

| Hierarchy level | Class | Number of videos | | | | Number of images (cumulative, including subclasses) |
|---|---|---|---|---|---|---|
| | | Year: 2021 | Year: 2023 | Total | Total, including subclasses | |
| | *Total* | 411 | 1,390 | 1,801 | 1,801 | 939,722 |
| L1 | *Insecta* | 2 | 31 | 33 | 1754 | 722,663 |
| | *Arachnida* | - | 14 | 14 | 14 | 2,832 |
| L2 | *Coleoptera* | - | - | - | 146 | 100,444 |
| | *Diptera* | 3 | 2 | 5 | 1,409 | 515,157 |
| | *Hymenoptera* | - | - | - | 90 | 53,905 |
| | *Lepidoptera* | 14 | 2 | 16 | 62 | 20,197 |
| | *Dermaptera* | - | 4 | 4 | 4 | 733 |
| | *Orthoptera* | 36 | 2 | 38 | 38 | 22,416 |
| | *Mecoptera* | 5 | - | 5 | 5 | 2,398 |
| L3 | *Apocrita* | 1 | - | 1 | 82 | 50,829 |
| | *Brachycera* | 108 | 393 | 501 | 1,365 | 501,679 |
| | *Nematocera* | 14 | 20 | 34 | 39 | 12,959 |
| | *Anthicidae* | 6 | - | 6 | 6 | 3,846 |
| | *Coccinelloidea* | 7 | 119 | 126 | 126 | 72,876 |
| | *Chrysomeloidea* | 2 | 2 | 4 | 4 | 10,645 |
| | *Heterocera* | 36 | 10 | 46 | 46 | 17,115 |
| | *Symphyta* | - | 8 | 8 | 8 | 3,076 |
| | *Curculionoidea* | 8 | 2 | 10 | 10 | 13,077 |
| L4 | *Aculeata* | 4 | 1 | 5 | 54 | 43,387 |
| | *Asilomorpha* | - | - | - | 81 | 26,516 |
| | *Muscomorpha* | 4 | 5 | 9 | 783 | 312,128 |
| | *Bibionidae* | - | 5 | 5 | 5 | 3,060 |
| | *Parasitica* | 19 | 8 | 27 | 27 | e7,442 |
| L5 | *Formicidae* | 16 | - | 16 | 16 | 40,064 |
| | *Empidoidea* | 21 | 1 | 22 | 22 | 2,418 |
| | *Oestroidea* | 10 | 57 | 67 | 67 | 15,958 |
| | *Ephydroidea* | 4 | 8 | 12 | 12 | 3,861 |
| | *Syrphidae* | 29 | 26 | 55 | 55 | 46,712 |
| | *Lauxanioidea* | 2 | 13 | 15 | 15 | 19,468 |
| | *Muscoidea* | 20 | 596 | 616 | 616 | 215,009 |
| | *Asiloidea* | 35 | 24 | 59 | 59 | 24,098 |
| | *Sepsidae* | 2 | 4 | 6 | 6 | 6,026 |
| | *Tephritidae* | - | 3 | 3 | 3 | 1,864 |
| | *Vespidae* | 3 | 30 | 33 | 33 | 3,323 |

*Table S2: Data splits showing the number of images per class in the training, validation and test splits and the relative percentages of final training images (Train+Validation) of each class, per level of hierarchy and for all samples in the data split.*

| Hierarchy level | Class | Data split (No. of images) | | | | % (Train+Val) of samples at hierarchy level | % (Train+Val) of total samples |
|---|---|---|---|---|---|---|---|
| | | Train | Validation | Retraining (Train+Val) | Test | | |
| L1 | *Insecta* | 201,828 | 61,157 | 262,985 | 75,555 | 98.93 | 98.93 |
| | *Arachnida* | 2,400 | 432 | 2,832 | 829 | 1.07 | 1.07 |
| L2 | *Coleoptera* | 34,224 | 11,344 | 45,568 | 9,551 | 17.83 | 17.14 |
| | *Diptera* | 104,208 | 32,627 | 136,835 | 40,544 | 53.54 | 51.48 |
| | *Hymenoptera* | 26,533 | 5,308 | 31,841 | 12,068 | 12.46 | 11.98 |
| | *Lepidoptera* | 14,133 | 6,064 | 20,197 | 4,903 | 7.90 | 7.60 |
| | *Dermaptera* | 557 | 176 | 733 | 285 | 0.29 | 0.28 |
| | *Orthoptera* | 15,445 | 2,555 | 18,000 | 5,734 | 7.04 | 6.77 |
| | *Mecoptera* | 1,766 | 632 | 2,398 | 720 | 0.94 | 0.90 |
| L3 | *Apocrita* | 24,072 | 4,693 | 28,765 | 9,486 | 12.46 | 10.82 |
| | *Brachycera* | 93,374 | 29,983 | 123,357 | 38,035 | 53.44 | 46.41 |
| | *Nematocera* | 10,367 | 2,592 | 12,959 | 2,130 | 5.61 | 4.88 |
| | *Anthicidae* | 3,077 | 769 | 3,846 | 178 | 1.67 | 1.45 |
| | *Coccinelloidea* | 13,846 | 4,154 | 18,000 | 6,000 | 7.80 | 6.77 |
| | *Chrysomeloidea* | 8,516 | 2,129 | 10,645 | 351 | 4.61 | 4.00 |
| | *Heterocera* | 11,658 | 5,457 | 17,115 | 4,262 | 7.41 | 6.44 |
| | *Symphyta* | 2,461 | 615 | 3,076 | 2,582 | 1.33 | 1.16 |
| | *Curculionoidea* | 8,785 | 4,292 | 13,077 | 3,022 | 5.66 | 4.92 |
| L4 | *Aculeata* | 17,486 | 3,837 | 21,323 | 7,018 | 15.54 | 8.02 |
| | *Asilomorpha* | 15,872 | 4,546 | 20,418 | 6,385 | 14.88 | 7.68 |
| | *Muscomorpha* | 63,641 | 21,298 | 84,939 | 25,650 | 61.92 | 31.95 |
| | *Bibionidae* | 2,448 | 612 | 3,060 | 1,155 | 2.23 | 1.15 |
| | *Parasitica* | 6,586 | 856 | 7,442 | 2,468 | 5.42 | 2.80 |
| L5 | *Formicidae* | 14,850 | 3,150 | 18,000 | 6,000 | 14.58 | 6.77 |
| | *Empidoidea* | 1,705 | 713 | 2,418 | 385 | 1.96 | 0.91 |
| | *Oestroidea* | 10,462 | 5,496 | 15,958 | 5,126 | 12.93 | 6.00 |
| | *Ephydroidea* | 2,586 | 1,275 | 3,861 | 716 | 3.13 | 1.45 |
| | *Syrphidae* | 13,914 | 4,086 | 18,000 | 6,000 | 14.58 | 6.77 |
| | *Lauxanioidea* | 13,742 | 4,258 | 18,000 | 5,591 | 14.58 | 6.77 |
| | *Muscoidea* | 14,082 | 3,918 | 18,000 | 6,000 | 14.58 | 6.77 |
| | *Asiloidea* | 14,167 | 3,833 | 18,000 | 6,000 | 14.58 | 6.77 |
| | *Sepsidae* | 4,821 | 1,205 | 6,026 | 510 | 4.88 | 2.27 |
| | *Tephritidae* | 1,491 | 373 | 1,864 | 1,031 | 1.51 | 0.70 |
| | *Vespidae* | 2,636 | 687 | 3,323 | 1,018 | 2.69 | 1.25 |